\documentclass{article}

\usepackage{arxiv}

\usepackage[utf8]{inputenc}
\usepackage[T1]{fontenc}
\usepackage{hyperref}
\usepackage{url}
\usepackage{booktabs}
\usepackage{amsfonts}
\usepackage{nicefrac}
\usepackage{microtype}
\usepackage{graphicx}
\usepackage{natbib}
\usepackage{doi}
\usepackage{tabularx}
\usepackage{makecell}
\usepackage{amsmath}
\usepackage{amssymb}
\usepackage{float}
\usepackage{subcaption}
\usepackage{threeparttable}
\usepackage{fancyhdr}
\date{}

\title{Sub-Metre Lunar DEM Generation and Validation from
Chandrayaan-2 OHRC Multi-View Imagery Using an Open-Source Pipeline}

\author{
	\href{https://orcid.org/0009-0000-3250-0753}{\includegraphics[scale=0.06]{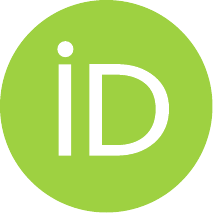}\hspace{1mm}Aaranay Aadi} \\
	School of Computer Science and Engineering\\
	Manipal University Jaipur\\
	Jaipur, India \\
	\texttt{aaranayaadi@gmail.com} \\
	\And
	\href{https://orcid.org/0000-0003-3893-2383}{\includegraphics[scale=0.06]{orcid.pdf}\hspace{1mm}Jai Singla} \\
	Space Applications Centre (SAC)\\
	Indian Space Research Organisation (ISRO)\\
	Ahmedabad, India \\
	\texttt{jaisingla@sac.isro.gov.in} \\
	\And
	\href{https://orcid.org/0000-0000-0000-0000}{\includegraphics[scale=0.06]{orcid.pdf}\hspace{1mm}Nitant Dube} \\
	Space Applications Centre (SAC)\\
	Indian Space Research Organisation (ISRO)\\
	Ahmedabad, India \\
	\texttt{nitant@sac.isro.gov.in} \\
    \And
	\href{https://orcid.org/0000-0001-7567-493X}{\includegraphics[scale=0.06]{orcid.pdf}\hspace{1mm}Oleg Alexandrov} \\
	Intelligent Robotics Group\\
	NASA Ames Research Center\\
	Moffet Field, CA, USA \\
	\texttt{oleg.alexandrov@nasa.gov} \\
}

\renewcommand{\shorttitle}{Sub-Metre Lunar DEM from Chandrayaan-2 OHRC}

\hypersetup{
  pdftitle={Sub-Metre Lunar DEM Generation and Validation from Chandrayaan-2 OHRC Multi-View Imagery Using Open-Source tools},
  pdfsubject={planetary science, photogrammetry, remote sensing},
  pdfauthor={Aaranay Aadi, Jai Singla, Nitant Dube, Oleg Alexandrov},
  pdfkeywords={Chandrayaan-2, OHRC, digital elevation model, photogrammetry, Ames Stereo Pipeline, Community Sensor Model, lunar terrain, stereo reconstruction},
}

\begin{document}
% Define the special style for the first page
\fancypagestyle{firstpage}{
  \fancyhf{} % clear all fields
  \renewcommand{\headrulewidth}{0pt} % remove horizontal line
  \fancyfoot[C]{\footnotesize This work is a preprint and has not been peer-reviewed. It has been posted to share preliminary findings. The views expressed here are those of the authors and do not necessarily reflect the views or policies of ISRO SAC.}
}

\pagestyle{empty} % Set the global style to empty (no footer on subsequent pages)

\maketitle
\thispagestyle{firstpage} % Apply the disclaimer style ONLY to this page
\pagestyle{empty}         % Ensures following pages remain empty}

\begin{abstract}
High-resolution digital elevation models (DEMs) of the lunar surface are essential for surface mobility planning, landing site characterization, and planetary science. The Orbiter High Resolution Camera (OHRC) on board Chandrayaan-2 has the best ground sampling capabilities of any lunar orbital imaging currently in use by acquiring panchromatic imagery at a resolution of roughly 20--30\,cm per pixel. This study presents a novel, reproducible framework for the generation of sub-metre DEMs from OHRC multi-view imagery using an exclusively open-source pipeline. Candidate stereo pairs are identified from non-paired OHRC archives through geometric analysis of image metadata, employing baseline-to-height (B/H) ratio computation and convergence angle estimation. Dense stereo correspondence and ray triangulation are then applied to generate point clouds, which are gridded into DEMs at effective spatial resolutions between approximately 24 and 54\,cm across five geographically distributed lunar sites. Geometric consistency is established through Iterative Closest Point (ICP) alignment against Lunar Reconnaissance Orbiter Narrow Angle Camera (NAC) Digital Terrain Models, followed by constant-bias offset correction. Post alignment comparison against NAC reference terrain yields a vertical RMSE of 5.85\,m (at native OHRC resolution), and a horizontal accuracy of within a pixel (approximately 30\,cm) assessed by planimetric feature matching.
\end{abstract}


% =============================================================================
\section{Introduction}
% =============================================================================

Digital Elevation Models (DEMs) of the lunar surface underpin a broad range of
scientific investigations, including geomorphological mapping, impact crater
morphometry, and slope hazard assessment for future robotic and human landing
missions. The accuracy and resolution of a DEM fundamentally determines the
scale of surface features that can be studied.

The primary source of high-resolution lunar topographic data has been the
Lunar Reconnaissance Orbiter (LRO), in particular the Lunar Orbiter Laser
Altimeter (LOLA) \citep{smith2010} and the Narrow Angle Camera (NAC)
\citep{robinson2010}. NAC stereo-derived DEMs achieve resolutions of
approximately 1--5\,m per pixel depending on imaging geometry and have proven
highly valuable for detailed terrain studies. At the planetary scale, the
SELENE Terrain Camera has also contributed global topographic models
\citep{araki2009}.

The Chandrayaan-2 Orbiter High Resolution Camera (OHRC) \citep{chowdhury2020}
represents a qualitative advance in ground sampling capability, acquiring
panchromatic imagery at approximately 25--30\,cm per pixel from a 100\,km
orbit \citep{isro2019}. This resolution, roughly an order of magnitude finer
than typical NAC DEMs, would allow DEM products capable of resolving
boulder-scale topographic features and metre-scale slope variations. Such data
are of direct relevance to landing site certification and rover traverse
planning for future lunar surface missions. The geometric fidelity of OHRC
imagery has previously been demonstrated for surface feature localisation,
including reconstruction of the Chandrayaan-3 rover traverse path
\citep{iyer2025}.

Despite this potential, the use of OHRC imagery for DEM generation within
open-source stereo pipelines remains limited. Two practical challenges have
contributed to this. First, OHRC images are not acquired or distributed as
explicit stereo pairs. As a result, suitable image combinations must be
identified from orbital geometry and acquisition metadata, rather than being
available a priori. Second, the Integrated Software for Imaging Spectrometers (ISIS)\citep{gaddis1997overview} and Ames Stereo Pipeline (ASP)\citep{shean2016,moratto2010,beyer2018,asp} are the most commonly used tools and OHRC data formats and sensor models are not
directly supported within them. Recent open-source efforts have
partially addressed this by enabling ingestion of OHRC data through
PDS4-compatible workflows and corresponding CSM-based camera models, allowing
stereo reconstruction to be carried out within a standard framework.

% =============================================================================
\section{Related Work}
% =============================================================================

\subsection*{From Laser Altimetry to High-Resolution Stereo Photogrammetry}
 
The history of lunar topographic mapping reflects a progressive shift from
global laser altimetry to dense regional photogrammetry, driven by the
increasing demand for terrain products at the scale of landing hazards.
Early global coverage was established by the SELENE Laser Altimeter (LALT)
and consolidated by LOLA, which provided a globally consistent vertical datum
at 30\,m horizontal resolution \citep{smith2010}. While both instruments offer
centimetre-level ranging precision, their across-track shot spacing, that is on the
order of 1--4\,km at the equator for LOLA, fundamentally limits their
utility for dense topographic mapping \citep{barker2016new}. The SLDEM2015
product, which fuses LOLA altimetry with SELENE Terrain Camera (TC) imagery,
achieves approximately 60\,m horizontal resolution \citep{barker2016new}, which
remains insufficient for decimeter-scale hazard identification.
 
Stereo photogrammetry fills this gap by exploiting image pairs acquired from
different viewing directions to reconstruct surface geometry at the spatial
resolution of the imaging sensor. The geometric principles underlying stereo
reconstruction, such as image matching, epipolar rectification, disparity
estimation, and ray triangulation, are well established
\citep{hartley2003}, and their application to planetary remote sensing has
been demonstrated across a wide range of missions and sensors. For Mars, the
combination of HiRISE (0.25\,m/pixel) and the Context Camera (CTX, 6\,m/pixel)
on the Mars Reconnaissance Orbiter has established a precedent for
nested-resolution mapping, in which CTX DEMs provide regional context for
HiRISE samples at the scale of boulders and slopes \citep{hepburn2019creating}. A
similar multi-resolution approach is now achievable on the Moon using LRO
NAC stereo at 1--2\,m and OHRC at 25--30\,cm, though the latter has not
previously been demonstrated.
 
\subsection*{Open-Source Pipeline}
 
The Ames Stereo Pipeline (ASP) \citep{moratto2010,shean2016,beyer2018} has
been the community standard for planetary stereo DEM production for over a
decade, providing an integrated suite of tools for radiometric normalisation,
feature matching, bundle adjustment, dense disparity estimation, ray
triangulation, and DEM gridding. ASP is built on the Vision Workbench library
and relies on ISIS \citep{gaddis1997overview} for planetary-specific data ingestion and
SPICE-based ephemeris and orientation (spiceinit) processing
\citep{stathopoulou2019}. Successful applications of this pipeline include
HiRISE stereo on Mars \citep{kirk2008}, LRO NAC stereo on the Moon
\citep{beyer2018}, and imagery from a growing range of international missions.
 
A significant recent development in the ASP ecosystem is the adoption of the
Community Sensor Model (CSM) \citep{csm2015} and the Abstraction Layer for
Ephemerides (ALE) \citep{laura2020planetary} as a more general mechanism for sensor
description. CSM-based camera models decouple the stereo pipeline from the
legacy ISIS internal camera models, enabling ASP to handle pushbroom sensors
from international missions, including those not originally designed with
ASP compatibility in mind, without requiring instrument-specific code within
the core pipeline \citep{liu2024}. This architectural shift is directly
relevant to the present work: the OHRC sensor model is implemented as a CSM
plugin, allowing standard ASP processing once the ISD file is generated via
ALE.
 
\subsection*{Lunar DEM Products and the Sub-Metre Gap}
 
The highest-resolution publicly available lunar DEMs are derived from LRO NAC
stereo pairs, typically at 1--2\,m post spacing \citep{beyer2018}. These
products have proven highly valuable for detailed crater morphometry and slope
mapping, but their resolution remains insufficient for identifying decimeter-scale
surface hazards such as rocks, hollows, and regolith texture variations that are
operationally relevant for landing site safety \citep{barker2016new}. Coverage
at the Lunar South Pole (LSP) is further constrained by the extreme illumination
geometry: near-polar imagery frequently contains large shadow regions that
frustrate stereo matching, and the near-nadir viewing geometry of LRO NAC at
high latitudes weakens the stereo convergence angle, reducing height sensitivity
\citep{liu2024}.
 
The OHRC instrument, with a GSD of 25--30\,cm, occupies the sub-metre niche
left unfilled by existing products. While LRO NAC offers broad coverage, its
resolution ceiling of approximately 0.5\,m per pixel (in the highest-resolution
mode, at roughly 50kms altitude) remains coarser than OHRC by a factor of roughly two, and the
strip-wise geometry of NAC acquisition makes it less well-suited to targeted
high-fidelity site surveys \citep{liu2024}.
 
\subsection*{Stereo Pair Identification from Non-Paired Archives}
 
Unlike dedicated stereo missions, which acquire
simultaneous multi-angle imagery in a single pass, OHRC operates in a
single-strip nadir-imaging mode. Viable stereo combinations must therefore
be identified retrospectively from the image archive using geometric metadata.
This ``opportunistic stereo'' approach has precedent in the context of
historical satellite imagery and for Mars HiRISE
\citep{kirk2008}, where overlapping strips acquired on different orbital passes
are evaluated for stereo suitability.
 
The primary geometric quality indicators for a candidate stereo pair are the
overlap fraction, the illumination similarity between acquisitions (quantified
by differences in solar incidence angle and azimuth), and the stereo strength
as expressed by the baseline-to-height (B/H) ratio \citep{becker2015criteria}. For
automated dense matching, a B/H ratio between 0.5 and 1.0 is generally
considered optimal: ratios below approximately 0.2 result in poor vertical
precision because the height error scales as $\Delta h \approx \text{GSD} /
(B/H)$, while ratios above approximately 1.2 introduce excessive perspective
distortion and parallax that causes systematic failure in dense matching
algorithms \citep{hasegawa2000accuracy,toutin2002}. Illumination consistency is a
further constraint specific to lunar stereo: because the Moon lacks an
atmosphere, surface brightness is dominated by solar incidence angle, and
pairs with large differences in solar geometry will exhibit photometric
disparities that degrade correspondence quality. Studies on PlanetScope
multi-view imagery have shown that even pairs with weak B/H ratios can yield
sub-metre vertical accuracy when image texture is high \citep{ghuffar2018},
but on the largely textureless highland regolith this condition does not
generally hold, making shadow-tip extent a practical additional selection
criterion \citep{becker2015criteria}.

\section{Contribution}
This work builds upon previous developments to enable and evaluate the
generation of high-resolution DEMs from Chandrayaan-2 OHRC stereo imagery
across multiple lunar regions using an open-source photogrammetric workflow.
The focus is on assessing the practical recoverability of sub-metre terrain
structure and quantifying the geometric and elevation consistency of the
resulting products with respect to reference NAC-derived datasets.

The contributions of this work are as follows:

\begin{itemize}
    \item Development of a reproducible open-source processing workflow for
          OHRC stereo imagery, including PDS4 ingestion support and CSM-based
          camera model configuration to enable compatibility with
          photogrammetric tools.

    \item Introduction of a systematic stereo pair selection strategy for
          non-catalogued OHRC image archives using baseline-to-height ratio
          and convergence angle constraints.

    \item Reconstruction of lunar DEMs from Chandrayaan-2 OHRC imagery at
          sub-metre spatial resolution (24--54\,cm), demonstrating preservation
          of fine-scale surface morphology across multiple lunar regions.

    \item Quantitative and qualitative evaluation of reconstructed DEMs
          against NAC-derived reference terrain for assessing geometric
          consistency and vertical accuracy.

    \item Post-processing of DEM products, including hole-filling and
          priority-based mosaicing, to improve spatial completeness and
          continuity of reconstructed regions.
\end{itemize}
% =============================================================================
\section{Dataset}
% =============================================================================

OHRC images used in this study were obtained from the Chandrayaan-2 Map Browse
facility of the Indian Space Research Organisation (ISRO). Each image is
distributed in PDS4 format, consisting of a binary image file accompanied by
an XML label containing instrument metadata, spacecraft state vectors, and
imaging geometry parameters.

OHRC raw (uncalibrated) data was used throughout this study in
preference to the calibrated data products also distributed by ISRO. This
choice was made for two reasons. First, ASP performs its own intensity
normalisation prior to stereo matching. Specifically, a per-image
normalisation to a mean~$\pm$~2\,$\sigma$ range, rather than a full
photometric calibration; since radiometric calibration typically applies a
scaling and a mildly non-linear response correction, the pipeline is
expected to perform comparably on either calibrated or uncalibrated input.
Second, the accuracy of the stereo-derived elevation model depends on the
geometric fidelity of the camera model, which is encoded in the SPICE
kernels\citep{acton1996} and CSM sensor description, rather than on radiometric correction;
using raw data therefore incurs no penalty in terms of the geometric quality
of the final DEM.

Spacecraft orientation and position data required to reconstruct the camera
geometry are provided through SPICE kernels distributed via
the ISRO Space Science Data Centre (ISSDC) PRADAN archive. These kernels encode
spacecraft ephemeris (SPK), spacecraft orientation (CK), instrument parameters
(IK), and planetary reference frames (PCK/FK) as a function of time.

Reference terrain data used for relative vertical consistency and hole-filling are
NAC-derived Digital Terrain Models from the LRO archive, available at 1--2\,m
resolution.

Table~\ref{tab:dataset} summarises the five stereo pairs used in this study,
including the product identifiers of each constituent image and their native
ground sampling distances as reported in the PDS4 XML metadata. Region~5,
which exhibits the highest B/H ratio and the most extreme convergence geometry
of the dataset, is discussed separately in Section~\ref{sec:discussion}.

\begin{table*}[h]
\centering
\small
\caption{OHRC image dataset: acquisition locations, product IDs, native ground
         sampling distances, and spacecraft altitudes. All Product IDs share the common prefix
         \texttt{ch2\_ohr\_nrp\_}.}
\label{tab:dataset}
\renewcommand{\arraystretch}{1.3}
\setlength{\tabcolsep}{5pt}
\newcolumntype{Y}{>{\centering\arraybackslash}X}
\begin{tabularx}{\textwidth}{c l Y Y c c c}
\toprule
\textbf{Region} &
\textbf{\makecell{Approx.\\ Location}} &
\textbf{\makecell{Product ID (Image 1)}} &
\textbf{\makecell{Product ID (Image 2)}} &
\textbf{\makecell{GSD\\ (cm/px)}} &
\textbf{\makecell{Alt$_1$\\ (km)}} &
\textbf{\makecell{Alt$_2$\\ (km)}} \\
\midrule
1 & $\sim\!69^{\circ}$S, $32^{\circ}$E
  & \texttt{20240425T1209509264} & \texttt{20240425T1406019344} & 0.26 & 101.90 & 102.35 \\
2 & $\sim\!5^{\circ}$N,  $234^{\circ}$E
  & \texttt{20220914T1033119094} & \texttt{20220914T0835371412} & 0.32 & 124.26 & 123.97 \\
3 & $\sim\!69^{\circ}$S, $342^{\circ}$E
  & \texttt{20230303T0350447888} & \texttt{20230303T0152168201} & 0.25 & 99.52 & 99.43 \\
4 & $\sim\!13^{\circ}$S, $25^{\circ}$E
  & \texttt{20210402T0155096873} & \texttt{20210401T2200364910} & 0.26 & 103.81 & 104.64 \\
5 & $\sim\!4^{\circ}$N,  $230^{\circ}$E
  & \texttt{20220321T0525226030} & \texttt{20220321T0326369085} & 0.19 & 76.66 & 76.32 \\
\bottomrule
\end{tabularx}

\vspace{4pt}
\end{table*}

% =============================================================================
\section{Methodology}
% =============================================================================

\begin{figure}[t]
\centering
\includegraphics[width=0.80\columnwidth]{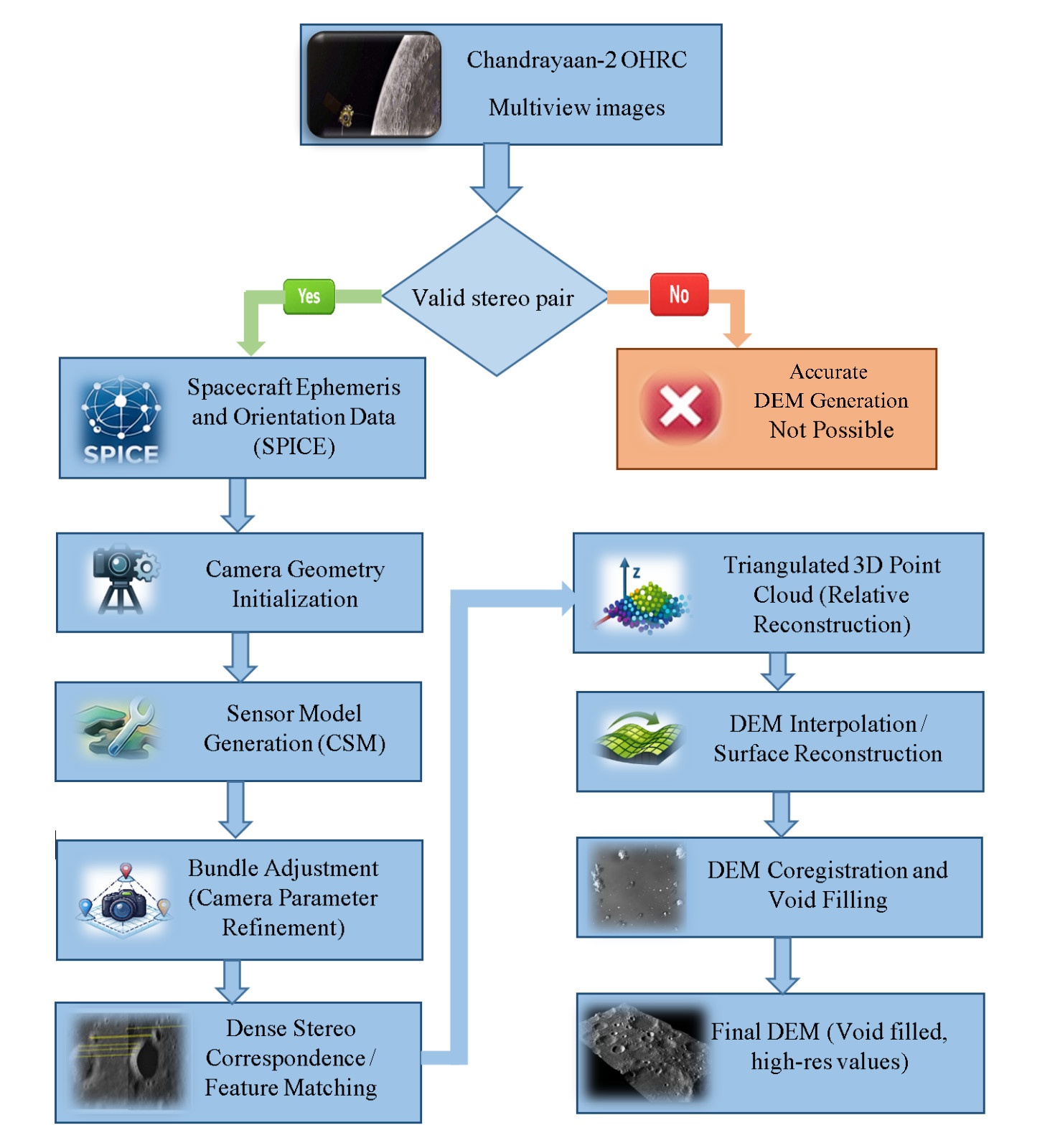}
\caption{Overview of the OHRC DEM generation pipeline. Stages proceed from
multi-view OHRC imagery and stereo pair selection through camera geometry
initialisation, sensor model generation (CSM), bundle adjustment (camera
parameter refinement), dense stereo correspondence and feature matching,
triangulated 3D point cloud reconstruction, DEM interpolation and surface
reconstruction, and finally DEM co-registration and void filling to yield the
final terrain product.}
\label{fig:workflow}
\end{figure}

The processing chain comprises six main stages: (1)~camera geometry
initialisation, (2)~stereo pair identification, (3)~bundle adjustment,
(4)~dense stereo reconstruction and point cloud generation, (5)~DEM gridding,
and (6)~post-processing including agreement with NAC reference terrain, offset correction, hole-filling, and mosaicing. Each stage is described below. The complete workflow
is illustrated in Fig.~\ref{fig:workflow}.

\subsection{OHRC Data Ingestion and SPICE Kernel Preparation}

Ingestion of OHRC data into ASP begins with the acquisition and preparation of
all relevant SPICE kernels from the ISSDC PRADAN archive. The kernel set
required for OHRC comprises the spacecraft trajectory kernel (SPK), the
attitude and pointing kernel (CK), the instrument kernel (IK) encoding the
OHRC focal geometry, and the planetary constants and frame kernels (PCK and
FK). Once downloaded, the kernels must be catalogued into an ISIS-format
database before they can be used for camera initialisation. This is
accomplished using \texttt{kerneldbgen}, an ISIS utility that parses each
kernel file and writes an indexed database file recording the kernel
type, coverage interval, and file path. ISIS subsequently queries these
database files by image acquisition time to select the correct kernels during
spice kernel initialization.

Native support for the Chandrayaan-2 OHRC instrument was not available in any
stable release of ISIS at the time this work began. Sensor model support for
OHRC was developed through active engagement with the open-source ISIS and ALE
communities, resulting in contributions to the ISIS development branch that
enabled correct ingestion and camera geometry initialisation for OHRC data
products. Initial DEM results were produced in November 2025 using ISIS
compiled from source against this development branch using CMake. Similarly, the ALE library needed to be compiled from source for generation of CSM camera models.

\subsection{Camera Geometry Initialisation and ASP Compatibility}

OHRC imagery is distributed in PDS4 format, which is not natively supported by
ASP's standard import routines. A dedicated PDS4 import template was developed
to map OHRC-specific metadata fields to ASP's internal ISIS cube format. This
template defines the instrument line and sample dimensions, exposure parameters,
and the mapping between XML label fields and ISIS keyword groups, enabling
ISIS to correctly ingest OHRC image products.

Following import, SPICE kernels are called on each ISIS cube to attach
spacecraft position and pointing information from the prepared SPICE database
to the image label. The camera projection for a surface point $\mathbf{P}$ in
the lunar body-fixed frame is then modelled as \citep{hartley2003}:
\begin{equation}
\mathbf{p} = K \begin{bmatrix} R & \mathbf{t} \end{bmatrix} \mathbf{P}
\label{eq:camera}
\end{equation}
where $K$ is the intrinsic calibration matrix encoding focal length and
principal point, $R$ is the rotation matrix derived from CK kernel data, and
$\mathbf{t}$ is the translation vector from SPK ephemeris data.

To enable open-source photogrammetric processing, camera models are generated
in the Community Sensor Model (CSM) format \citep{csm2015} using the ALE
(Abstraction Layer for Ephemerides) library \citep{laura2020planetary}. This utility
reads the SPICE-initialised ISIS cube and produces a JSON-format Instrument
Support Data (ISD) file describing the complete sensor geometry required by
ASP. Enabling this pipeline for OHRC required the development of a CSM sensor
model configuration not previously available for this instrument.

\subsection{Stereo Pair Identification}

Because OHRC images are not acquired or catalogued as stereo pairs, viable
stereo combinations must be identified from scratch using geometric parameters
extracted from image XML metadata. Two images of the same surface are considered
a candidate stereo pair if their relative viewing geometry satisfies constraints
on the baseline-to-height ratio and the convergence angle.

The baseline $B$ between two acquisitions is the distance between the
spacecraft positions $\mathbf{S}_1$ and $\mathbf{S}_2$ at the time of each
image:
\begin{equation}
B = \| \mathbf{S}_1 - \mathbf{S}_2 \|
\end{equation}

The baseline-to-height ratio is then
\begin{equation}
\frac{B}{H} = \frac{\| \mathbf{S}_1 - \mathbf{S}_2 \|}{H}
\end{equation}
where $H$ is the spacecraft altitude above the surface. The B/H ratio is a
fundamental descriptor of stereo geometry, controlling the trade-off between
height sensitivity and image similarity. Low B/H values reduce vertical
precision, whereas high values increase geometric and radiometric disparities,
making feature matching less reliable. In practice, moderate B/H values are
generally preferred for robust DEM generation \citep{bervoets1977elements,jing2008research, esri_bh}.

The convergence angle $\theta$ between the two viewing directions
$\hat{\mathbf{v}}_1$ and $\hat{\mathbf{v}}_2$ is
\begin{equation}
\theta = \cos^{-1}(\hat{\mathbf{v}}_1 \cdot \hat{\mathbf{v}}_2)
\end{equation}

For a pushbroom sensor observed from altitude $H$ with baseline $B$, the
convergence angle is related to the B/H ratio approximately as
\begin{equation}
\theta \approx 2 \arctan\!\left(\frac{B/H}{2}\right)
\label{eq:conv_approx}
\end{equation}

This approximation follows from basic geometric considerations because of a symmetric
viewing configuration ~\citep{hartley2003}. It was used to estimate convergence angles for
all candidate pairs in this study. The expected height precision (EP) of the resulting DEM is inversely related to the stereo convergence angle and can be approximated (from standard stereo error propagation principles) as
\begin{equation}
\mathrm{EP} \propto \frac{\rho \cdot \mathrm{GSD}}{\tan \theta}
\end{equation}
where $\rho$ is the image matching precision in pixels and GSD is the ground sampling distance.

Candidate pairs were identified by searching for OHRC images with overlapping
ground footprints acquired within temporal windows short enough to ensure
consistent surface illumination. B/H ratios and convergence angles were
computed for all footprint-overlapping pairs and filtered against the geometric
thresholds described above. The five stereo pairs selected for this study and
their stereo geometry parameters are listed in Table~\ref{tab:results}.

\subsection{Bundle Adjustment}

Even with accurately initialised SPICE-based camera models, residual pointing
errors between stereo images introduce systematic offsets that degrade
reconstruction accuracy. Bundle adjustment (~\cite{triggs1999bundle}) jointly refines the camera
parameters and the positions of matched feature points to minimise overall
geometric inconsistency.
 
The objective minimised during bundle adjustment is the reprojection error
summed over all matched tie points, wrapped in a robust cost function to
attenuate the influence of outliers:
\begin{equation}
E = \sum_{i=1}^{N} \rho\!\left(\left\| \mathbf{p}_i -
    \hat{\mathbf{p}}_i(\mathbf{C}, \mathbf{X}_i) \right\|^2\right)
\end{equation}
where $\mathbf{p}_i \in \mathbb{R}^2$ is the observed image coordinate of tie
point $i$, and $\hat{\mathbf{p}}_i$ is its reprojection given the current
camera parameters $\mathbf{C}$ and 3D point position $\mathbf{X}_i$. The
function $\rho(\cdot)$ is a robust Cauchy loss,
\begin{equation}
\rho(s) = c^2 \log\left(1 + \frac{s}{c^2}\right),
\end{equation}
where $c$ is a scale parameter controlling the influence of large residuals.
This formulation down-weights outliers arising from mismatched tie points and
improves the stability of the bundle adjustment solution. Optimisation was carried out using the Levenberg--Marquardt solver
(implemented in the Google Ceres library \citep{agarwal2022}) over up to 100 iterations. In addition to
the reprojection term, ASP's bundle adjustment incorporates a ground
constraint based on triangulated point positions to prevent the camera
solution from drifting away from the initial SPICE geometry, and optionally
allows the user to specify a camera position uncertainty constraint. The
refined camera states are saved and used as the adjusted pointing model for
all subsequent stereo processing.

\subsection{Dense Stereo Reconstruction}
Following bundle adjustment, dense stereo matching is performed to compute
per-pixel disparities between the rectified image pair. The disparity field
$\mathbf{d}(u,v)$ encodes the horizontal displacement between corresponding
pixels in the two images. The default stereo algorithm in ASP is a
block-matching approach based on normalised cross-correlation, which was used
in this study. 

While ASP provides the More Global Matching (MGM) method of
\citet{facciolo2015} as a more advanced alternative, block matching was found
to be sufficient for OHRC imagery due to its high spatial resolution and
strong local contrast, which provide reliable matching cues at the native
scale. Additionally, the lower computational and memory overhead of block
matching makes it well-suited for processing large OHRC scenes without
compromising reconstruction quality in well-textured regions.

Each matched pixel pair $(\mathbf{p}_1, \mathbf{p}_2)$ defines a pair of
viewing rays from cameras $C_1$ and $C_2$. The 3D position of the
corresponding surface point is determined by ray triangulation:
\begin{equation}
\mathbf{X} = \operatorname{Triangulate}(C_1, C_2, \mathbf{p}_1, \mathbf{p}_2)
\end{equation}
where $C_1$ and $C_2$ are the refined camera projection matrices from bundle
adjustment and $\mathbf{p}_1$, $\mathbf{p}_2$ are the matched image
coordinates. In practice the two rays are not exactly coplanar, so
$\mathbf{X}$ is computed as the midpoint of the shortest segment connecting the
two rays. Applying this over all matched pixels produces a dense 3D point cloud
representing the lunar surface.

\subsection{DEM Gridding}

The irregular 3D point cloud is rasterised onto a regular geographic grid to
produce the DEM. For each grid cell at $(x, y)$, the elevation is estimated by
interpolating over neighbouring point cloud samples:
\begin{equation}
z(x, y) = \frac{\displaystyle\sum_{i} w_i\, z_i}
               {\displaystyle\sum_{i} w_i}
\end{equation}
where $z_i$ are the elevations of points within the interpolation kernel and
$w_i$ are distance-based weights. Grid cells for which no point cloud samples
fall within the search radius are marked as no-data and subsequently addressed
in the hole-filling stage. DEM gridding was performed using \texttt{point2dem}
within ASP, with the output projected in a stereographic projection appropriate
to the region of interest.

\subsection{Alignment with NAC Reference}

The DEM produced from stereo reconstruction represents elevations in a local reference frame and may exhibit a global rigid-body offset relative to the lunar body-fixed coordinate system due to residual camera geometry and orbit/attitude uncertainties. To address this, a rigid co-registration is performed using an Iterative Closest Point (ICP) approach \citep{besl1992}, aligning the reconstructed OHRC point cloud to a co-located Lunar Reconnaissance Orbiter (LRO) NAC Digital Terrain Model (DTM) serving as the reference surface.

Point-to-surface ICP (projective ICP) estimates the rigid transformation
\begin{equation}
\mathbf{P}^{\prime} = R\,\mathbf{P} + \mathbf{T}
\end{equation}
that minimises the point-to-surface error between the transformed OHRC points and the NAC reference terrain:
\begin{equation}
\underset{R,\,\mathbf{T}}{\arg\min} \;
\frac{1}{N}\sum_{i=1}^{N}
\left\| \mathbf{P}^{\prime}_i - \Pi_{\text{NAC}}(\mathbf{P}^{\prime}_i) \right\|^2
\end{equation}
where $\Pi_{\text{NAC}}(\cdot)$ denotes the orthogonal projection of a point onto the NAC surface, and $R \in SO(3)$, $\mathbf{T} \in \mathbb{R}^3$. The resulting transformation is applied to the OHRC point cloud prior to final DEM interpolation, ensuring geometric consistency with the NAC reference frame.

\subsection{DEM Offset Correction}

Following ICP alignment, terrain profile comparison between the OHRC DEM and
the NAC DTM may reveal a residual constant vertical bias. The primary source
of this offset is inaccuracy in the SPICE kernels distributed with the
Chandrayaan-2 mission: errors in the spacecraft ephemeris (SPK) or attitude
(CK) kernels propagate directly into the reconstructed camera positions and
therefore into the absolute elevation of the derived point cloud. While ICP
alignment reduces the bulk of this offset by registering the point cloud
rigidly to the NAC reference surface, a residual constant bias may persist
if the kernel error manifests primarily as a uniform altitude shift rather
than a tilt or rotation. The corrected elevation is
\begin{equation}
z_{\mathrm{corrected}}(x,y) = z_{\mathrm{DEM}}(x,y) - \Delta z
\end{equation}
where $\Delta z$ is the mean vertical offset estimated from profile analysis
along representative terrain transects (using QGIS). Only constant-offset
corrections (i.e., uniform vertical shifts with no tilt component) are applied;
tilt residuals, if present, are addressed by the ICP step.

\subsection{Hole Filling Using NAC Terrain Data}

Stereo reconstruction at sub-metre resolution produces no-data regions wherever
image matching fails, typically in shadow-affected areas or surfaces with
insufficient photometric texture. Often times the images exhibit illumination inconsistencies, which adversely affect feature matching and result in void fractions in the reconstructed DEM. To produce a continuous terrain model, missing elevations are filled using the
aligned NAC DTM. Let $D_{\mathrm{OHRC}}(x,y)$ denote the OHRC-derived elevation
and $D_{\mathrm{NAC}}(x,y)$ the reference elevation. The merged elevation is
defined as
\begin{equation}
D(x,y) =
\begin{cases}
D_{\mathrm{OHRC}}(x,y), & \text{if } D_{\mathrm{OHRC}}(x,y) \text{ is valid} \\
D_{\mathrm{NAC}}(x,y),  & \text{otherwise}
\end{cases}
\end{equation}
where validity is determined by the presence of a reconstructed elevation
(i.e., non-void pixels in the OHRC DEM).

\subsection{Priority-Based DEM Mosaicing}

Where multiple DEM tiles from adjacent or overlapping stereo pairs cover a
common region, the tiles are combined into a single continuous product. A
priority-based blending strategy is applied using the \texttt{dem\_mosaic} tool
(ASP), which assigns higher priority to OHRC-derived elevations over NAC terrain
within a configurable blending-length transition zone.

The final merged elevation at $(x,y)$ is given by
\begin{equation}
D_{\mathrm{final}}(x,y) = \alpha(x,y)\, D_{\mathrm{OHRC}}(x,y)
  + \bigl(1 - \alpha(x,y)\bigr)\, D_{\mathrm{NAC}}(x,y)
\end{equation}
where $\alpha(x,y) \in [0,1]$ is a spatially varying weight that equals 1 in
regions with full OHRC coverage and transitions smoothly to 0 at the boundaries
of the OHRC footprint, ensuring seamless blending. Such weighted feathering is
commonly used in DEM and image mosaicing workflows \citep{asp}.

Within OHRC-covered regions $\alpha = 1$, and the formulation reduces to the
hole-filling scheme described above. The blending transition length was set to
14 pixels.

% =============================================================================
\section{Results}
\label{sec:results}
% =============================================================================

The proposed workflow was applied to five lunar regions spanning a range of
latitudes and terrain types. Table~\ref{tab:results} summarises the computed
B/H ratios, convergence angles, and the spatial resolutions of the generated
DEMs. Convergence angles were computed from B/H ratios using
Eq.~\eqref{eq:conv_approx}; for Region~5 the directly measured angle of
$\sim\!61^{\circ}$ is reported, as the approximation underestimates the true
angle at high B/H values.

\begin{table*}[t]
\centering
\begin{threeparttable}
\caption{Stereo geometry and generated DEM characteristics for each region,
          ordered by increasing B/H ratio.}
\label{tab:results}
\renewcommand{\arraystretch}{1.3}
\begin{tabular}{clccc}
\toprule
\textbf{Region} & \textbf{Approx.\ Location} & \textbf{B/H Ratio} &
\textbf{Conv.\ Angle ($^{\circ}$)} & \textbf{DEM Resolution} \\
\midrule
1 & $\sim\!69^{\circ}$S, $32^{\circ}$E    & 0.396 & $\sim\!22^{\circ}$ & 31\,cm \\
2 & $\sim\!5^{\circ}$N,  $234^{\circ}$E  & 0.414 & $\sim\!23^{\circ}$ & 54\,cm \\
3 & $\sim\!69^{\circ}$S, $342^{\circ}$E  & 0.559 & $\sim\!31^{\circ}$ & 30\,cm \\
4 & $\sim\!13^{\circ}$S, $25^{\circ}$E   & 0.877 & $\sim\!47^{\circ}$ & 48\,cm \\
\textbf{\textit{5}} & \textbf{\textit{$\sim\!4^{\circ}$N, $230^{\circ}$E}} & \textbf{\textit{1.161}}$^{\ddagger}$ & \textbf{\textit{$\sim\!61^{\circ}$}}$^{\dagger}$ & \textbf{\textit{24\,cm}} \\
\bottomrule
\end{tabular}
\begin{tablenotes}[para,flushleft]
    \small
    \item $^{\dagger}$ Directly measured; the approximation of
          Eq.~\eqref{eq:conv_approx} underestimates $\theta$ at high B/H.
    \item $^{\ddagger}$ High void fraction; see Section~\ref{sec:discussion}
          for details regarding performance at high B/H.
\end{tablenotes}
\end{threeparttable}
\end{table*}

B/H ratios across the five pairs range from 0.396 (Region~1) to 1.161
(Region~5), corresponding to convergence angles of approximately $22^{\circ}$
to $61^{\circ}$. Regions~1--4 all fall within the commonly used B/H
range of 0.3--1.0 \citep{hasegawa2000accuracy} and produced DEMs with satisfactory
spatial completeness.

Completed DEMs for Regions~1--4 achieve effective spatial resolutions between
31\,cm and 54\,cm, representing the highest-resolution DEM products reported
from Chandrayaan-2 OHRC data to date and covering four geographically
distributed sites. Region~5, while yielding the finest nominal grid spacing
(24\,cm), exhibits a substantially elevated void fraction and is discussed as a
limiting case in Section~\ref{sec:discussion}. The variation in resolution
across regions reflects differences in OHRC image sampling at acquisition time.
Alignment with NAC DTMs established absolute elevation consistency, and
priority-based mosaicing produced continuous terrain products.

\begin{figure}[h]
\centering
\includegraphics[width=0.95\columnwidth, height=6cm, keepaspectratio]{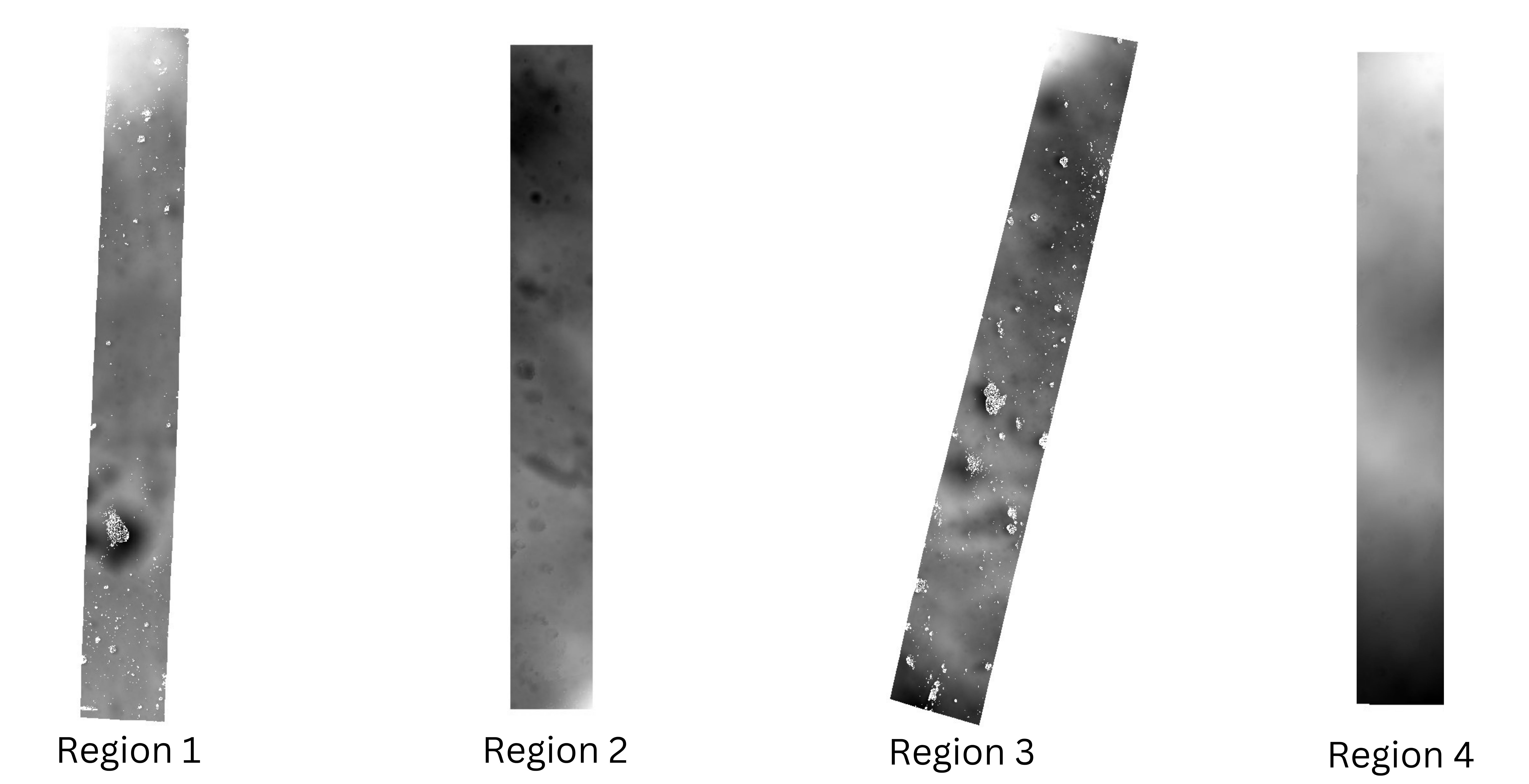}
\caption{DEMs for four of the reconstructed regions. Region~5 is demonstrated in section~\ref{sec:discussion} for its high void fraction. Regions~2 and 4
have well-filled DEMs with minimal voids. Region~1 was not fully filled using
NAC due to the presence of holes in the corresponding NAC DTM. Region~3 has no
corresponding NAC reference available.}
\label{fig:dem_all}
\end{figure}

The five DEMs are shown together in Fig.~\ref{fig:dem_all}. Regions~2 and ~4
exhibit well-filled terrain models with minimal void fraction. Region~1 retains
some unfilled areas owing to gaps in the corresponding NAC DTM used for infill.
Region~3 had no NAC reference available and is therefore presented without void
filling.

\begin{figure}[H]
     \centering
     \begin{subfigure}[b]{0.49\textwidth}
         \centering
         \includegraphics[width=\textwidth]{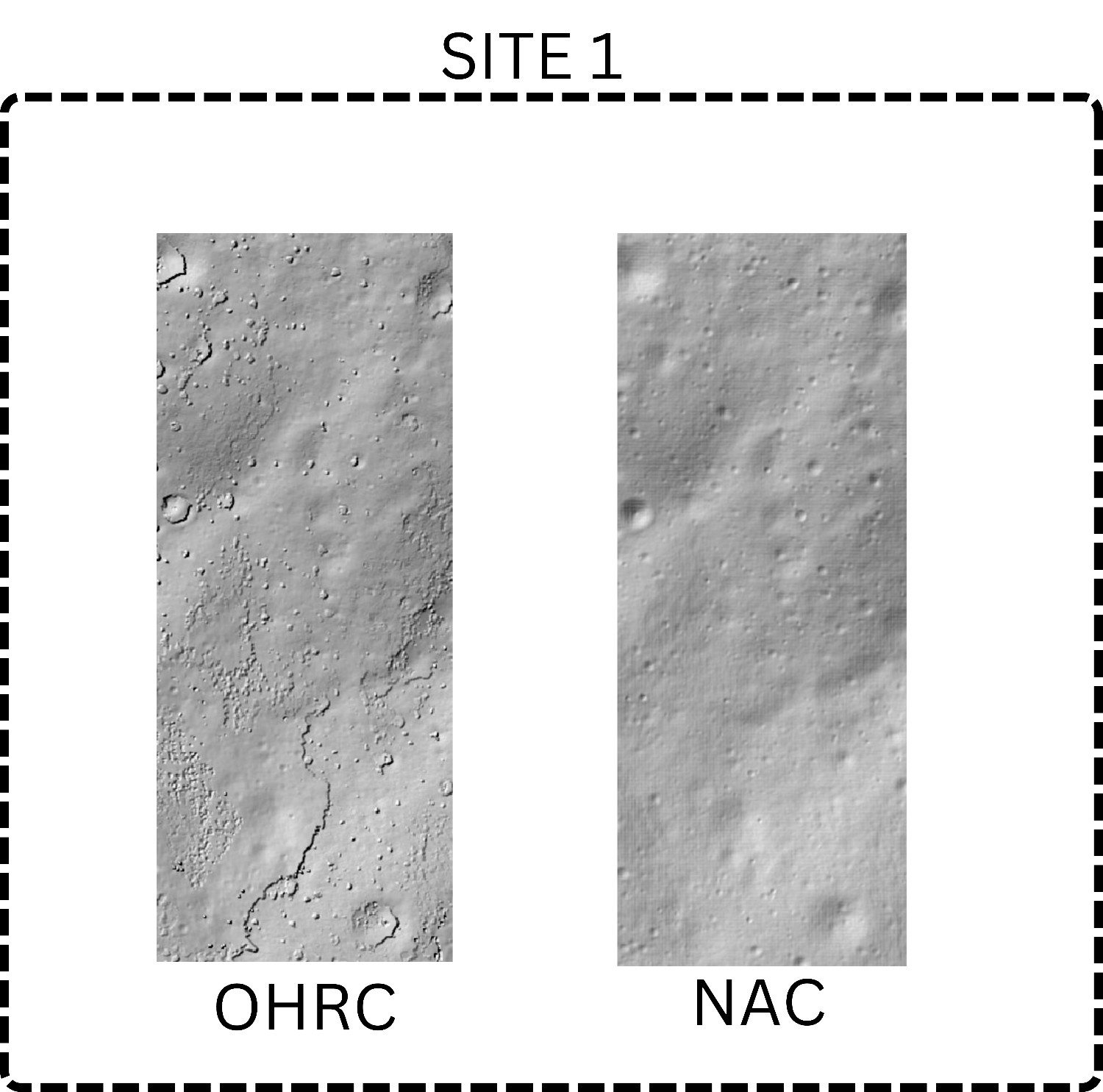}
         \caption{Site 1}
         \label{fig:site1}
     \end{subfigure}
     \hfill
     \begin{subfigure}[b]{0.49\textwidth}
         \centering
         \includegraphics[width=\textwidth]{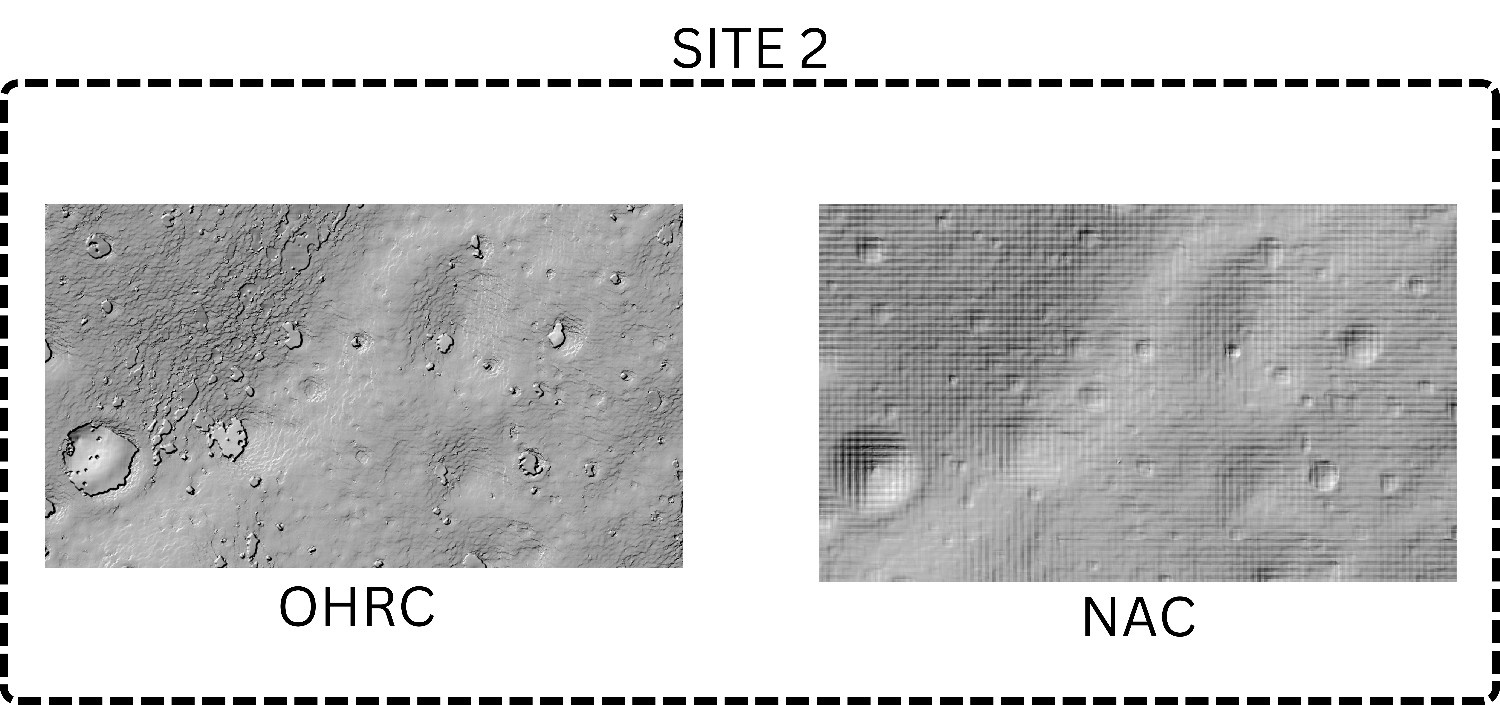}
         \caption{Site 2}
         \label{fig:site2}
     \end{subfigure}
     \caption{Visual feature comparison of OHRC generated DEM vs NAC DTM
     (both hillshaded) for Sites~1 and 2, randomly selected from Region~4.}
     \label{fig:comparison_1_2}
\end{figure}

\begin{figure}[H]
     \centering
     \begin{subfigure}[b]{0.49\textwidth}
         \centering
         \includegraphics[width=\textwidth]{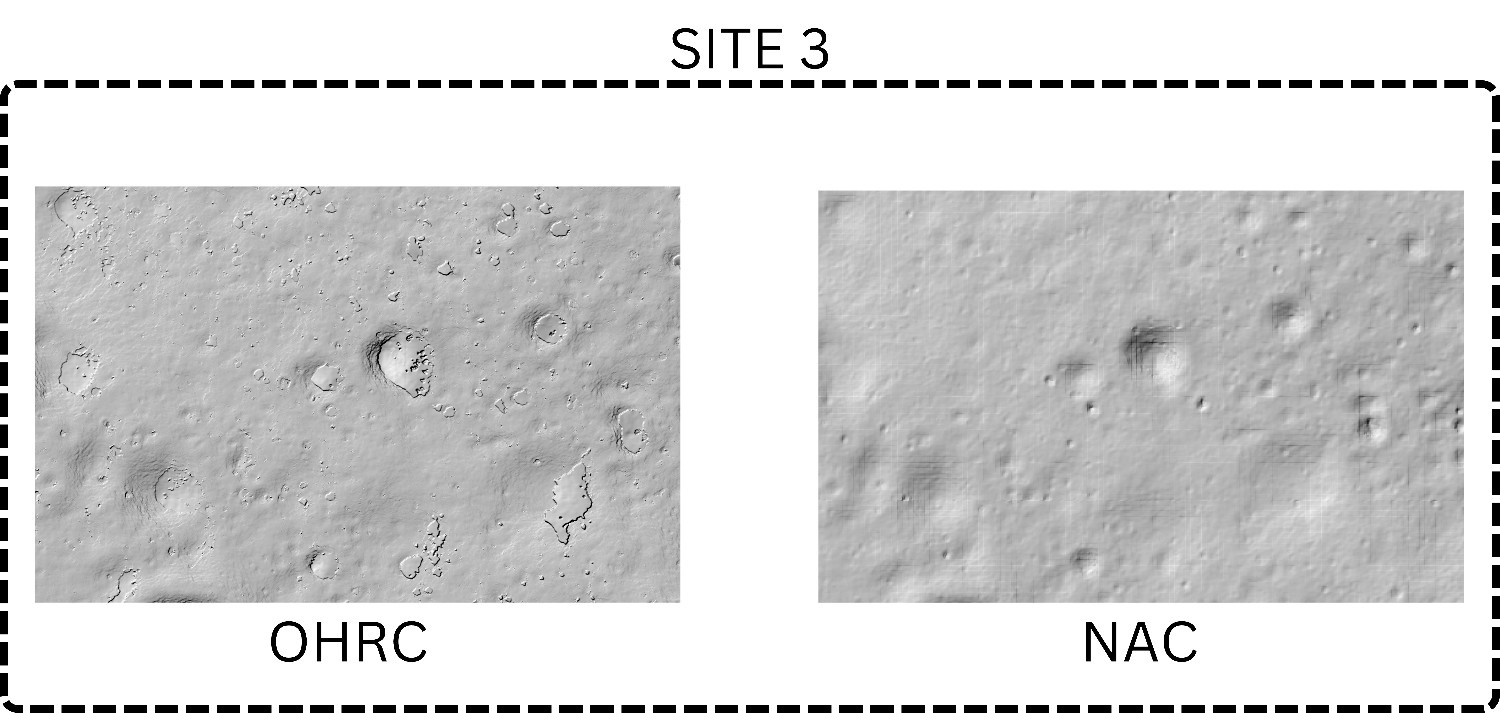}
         \caption{Site 3}
         \label{fig:site3}
     \end{subfigure}
     \hfill
     \begin{subfigure}[b]{0.49\textwidth}
         \centering
         \includegraphics[width=\textwidth]{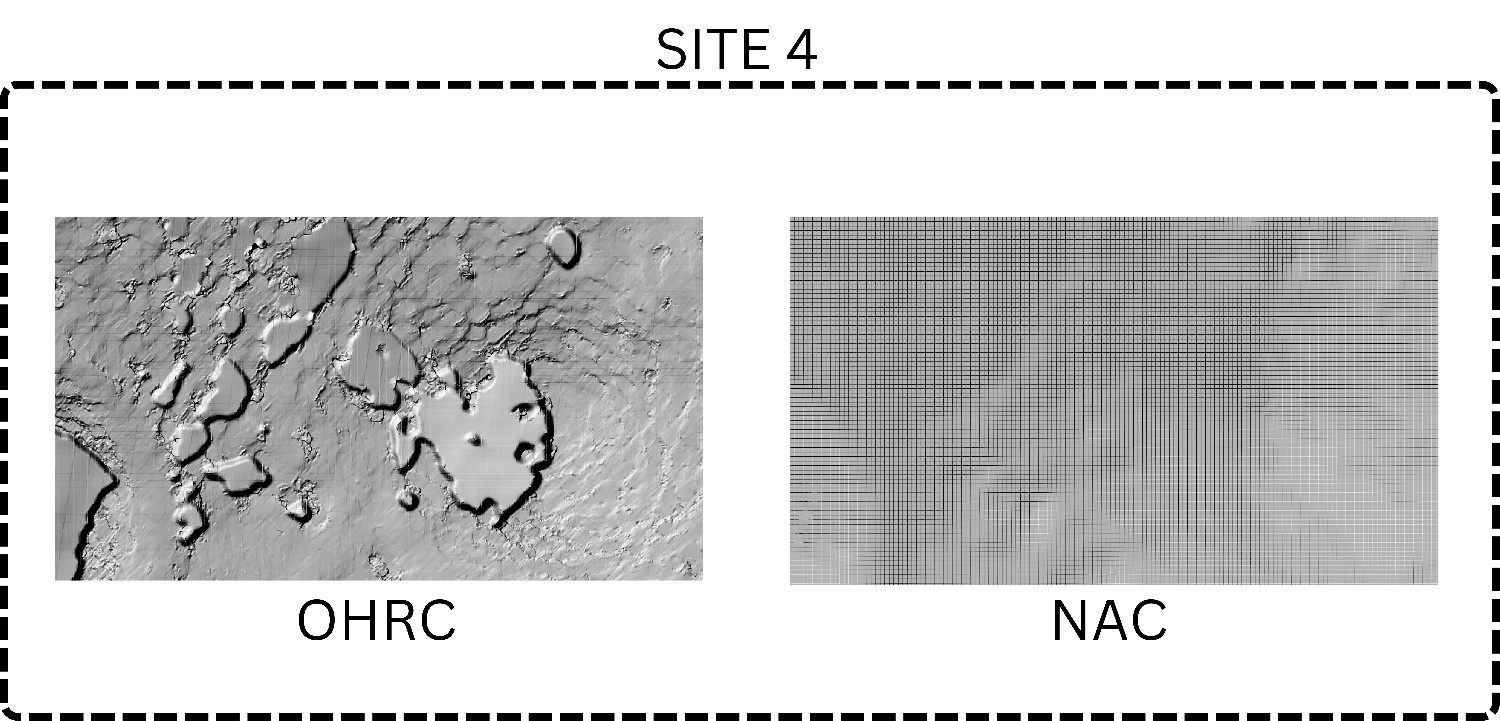}
         \caption{Site 4}
         \label{fig:site4}
     \end{subfigure}
     \caption{Visual feature comparison of OHRC generated DEM vs NAC DTM
     (both hillshaded) for Sites~3 and 4, randomly selected from Region~2.}
     \label{fig:comparison_3_4}
\end{figure}

Figs.~\ref{fig:comparison_1_2} and \ref{fig:comparison_3_4} present
side-by-side hillshade comparisons of the OHRC-derived DEM and the NAC DTM at
four randomly selected sites drawn from Regions~1 and 4 respectively,
illustrating the close visual agreement in terrain morphology between the two
products at sub-metre scales.

\subsection{Validation Against NAC Reference Terrain}

Metrological validation of the generated DEMs was carried out by comparing
elevation profiles extracted from the OHRC DEM against co-located profiles from
the NAC DTM reference. Profile transects were drawn across representative
terrain features using
QGIS, including crater rims and flat mare surfaces, ensuring that the comparison spans both high-relief and low-relief
terrain.

Let $z_{\mathrm{OHRC}}(s)$ and $z_{\mathrm{NAC}}(s)$ denote the elevation
values along a profile parameterised by arc length $s$. The vertical
discrepancy at each sample is
\begin{equation}
\delta(s) = z_{\mathrm{OHRC}}(s) - z_{\mathrm{NAC}}(s)
\end{equation}

The root-mean-square error (RMSE) of the vertical difference over the profile
is
\begin{equation}
\mathrm{RMSE}_v = \sqrt{\frac{1}{N}\sum_{i=1}^{N} \delta(s_i)^2}
\end{equation}

Across the validated regions (Regions~1, 2, and 4), the vertical RMSE values
are reported in Table~\ref{tab:rmse}. It should be noted that these figures
reflect a comparison between the OHRC DEM at its native sub-metre resolution
and the NAC DTM at approximately 3\,m posting.
An exactly similar vertical accuracy assessment, in which the OHRC DEM was
downsampled to NAC resolution prior to differencing, was also conducted, yielding roughly the same values. This confirms that the obtained results are reproducible and valid for lunar applications. Region~3 does not have a corresponding NAC reference and could not be
evaluated. Region~5 was excluded from the primary validation owing to its
elevated void fraction.

\begin{table}[h]
\centering
\begin{threeparttable}
\caption{Vertical RMSE by region (computed from the standard deviation of
         elevation residuals after bias correction, compared against the
         NAC DTM).}
\label{tab:rmse}
\begin{tabular}{lc}
\toprule
\textbf{Region} & \textbf{RMSE (cm)} \\
\midrule
1 & 298.96 \\
2 & 850.42 \\
3 &  --- (no NAC DTM available)\\
4 & 606.66 \\
5 & --- (excluded; high void fraction) \\
\midrule
\textbf{Mean (Regions~1, 2, 4)} & \textbf{585.35\,$\approx$\,5.85\,m} \\
\bottomrule
\end{tabular}
\end{threeparttable}
\end{table}

Horizontal accuracy was assessed independently of the vertical comparison by
identifying morphologically distinct surface features, including small crater
rims and sharp ridge crests, that are unambiguously localised in both the
OHRC DEM hillshade and the NAC DTM hillshade. Between 5 and 10 such features
were identified per validated region, and their planimetric centroids were
measured in QGIS. The mean horizontal offset across all matched features was
found to be less than 30\,cm, consistent with the native GSD of the OHRC
instrument and with horizontal accuracy bounds reported for analogous
high-resolution pushbroom stereo products. 

Horizontal accuracy was further corroborated by the mean triangulation error across Regions~1--4, computed from the intersection distance between back-projected stereo rays. The mean triangulation error was $0.212 \pm 0.164$\,m (pooled mean $\pm$ mean standard deviation across the four regions), consistent with the sub-30\,cm horizontal accuracy observed through planimetric feature matching.

\begin{figure}[H]
\centering
\includegraphics[width=0.95\columnwidth]{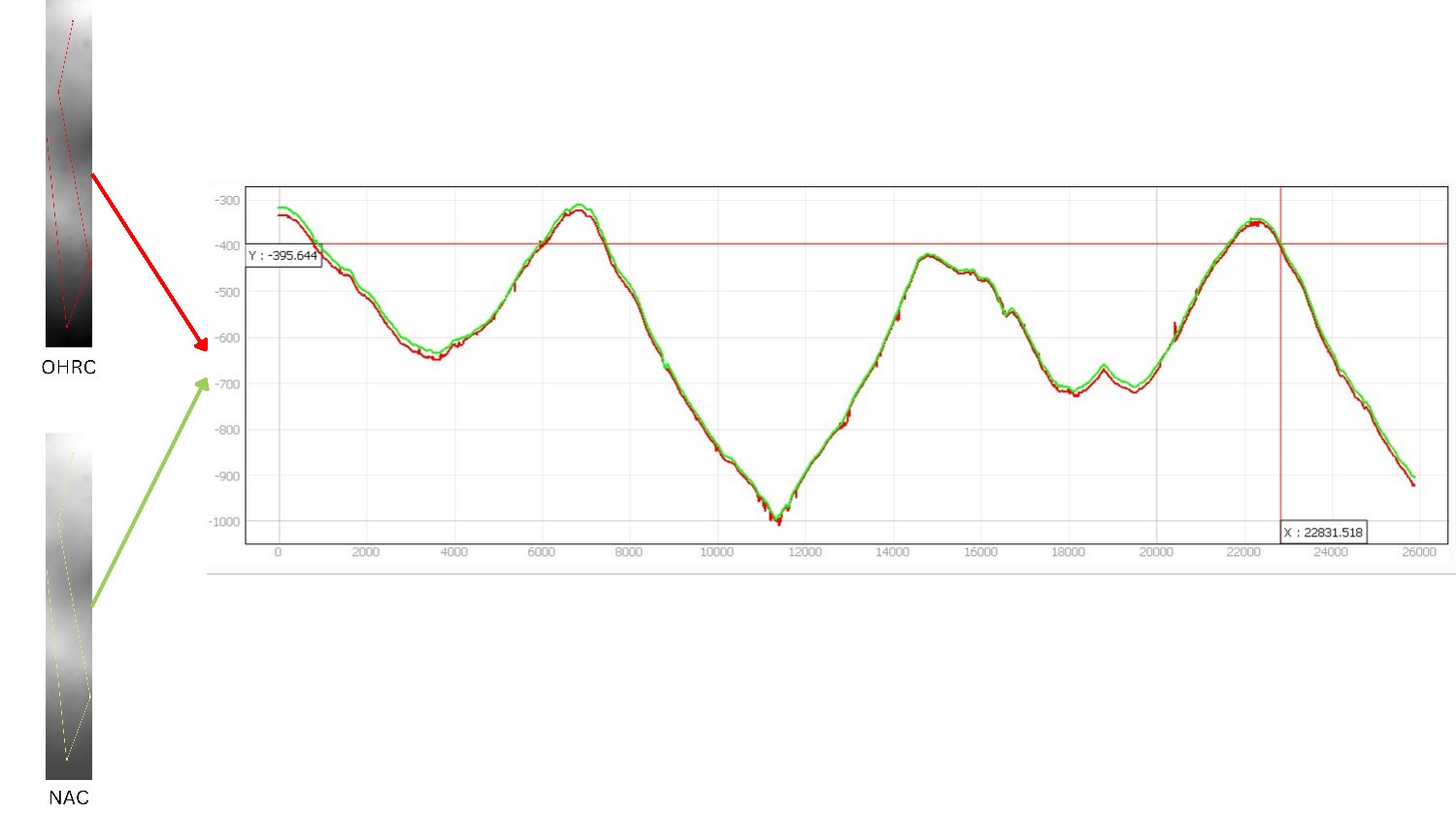}
\caption{Terrain profile comparison across Region 4. Horizontal axis: distance along profile
transect (m). Vertical axis: elevation (m). Red curve: OHRC DEM. Green curve:
NAC DTM reference.}
\label{fig:validation1}
\end{figure}

\begin{figure}[H]
\centering
\includegraphics[width=0.95\columnwidth]{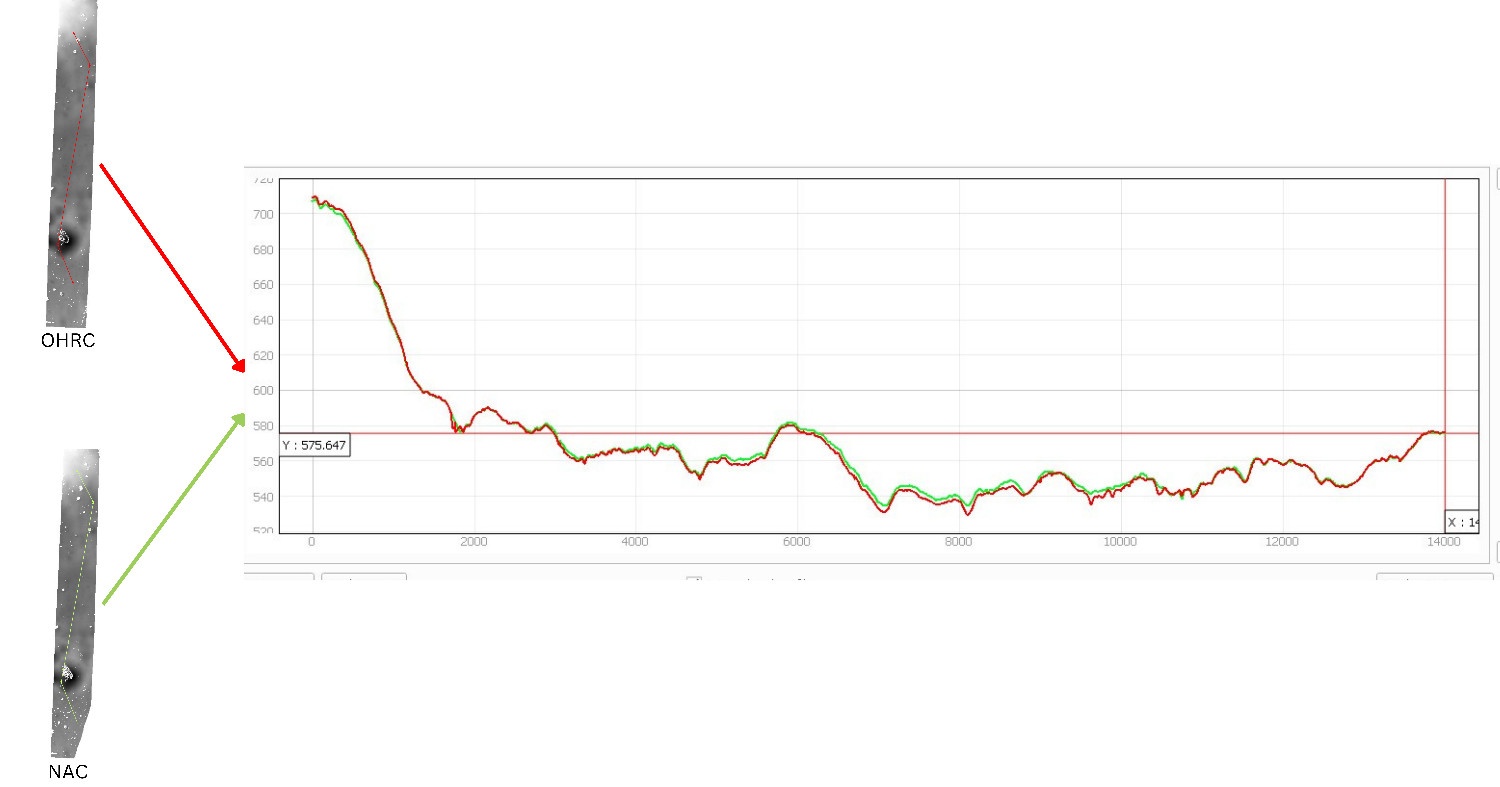}
\caption{Terrain profile comparison across Region 1. Horizontal axis: distance along profile
transect (m). Vertical axis: elevation (m). Red curve: OHRC DEM. Green curve:
NAC DTM reference.}
\label{fig:validation2}
\end{figure}

\begin{figure}[H]
\centering
\includegraphics[width=0.95\columnwidth]{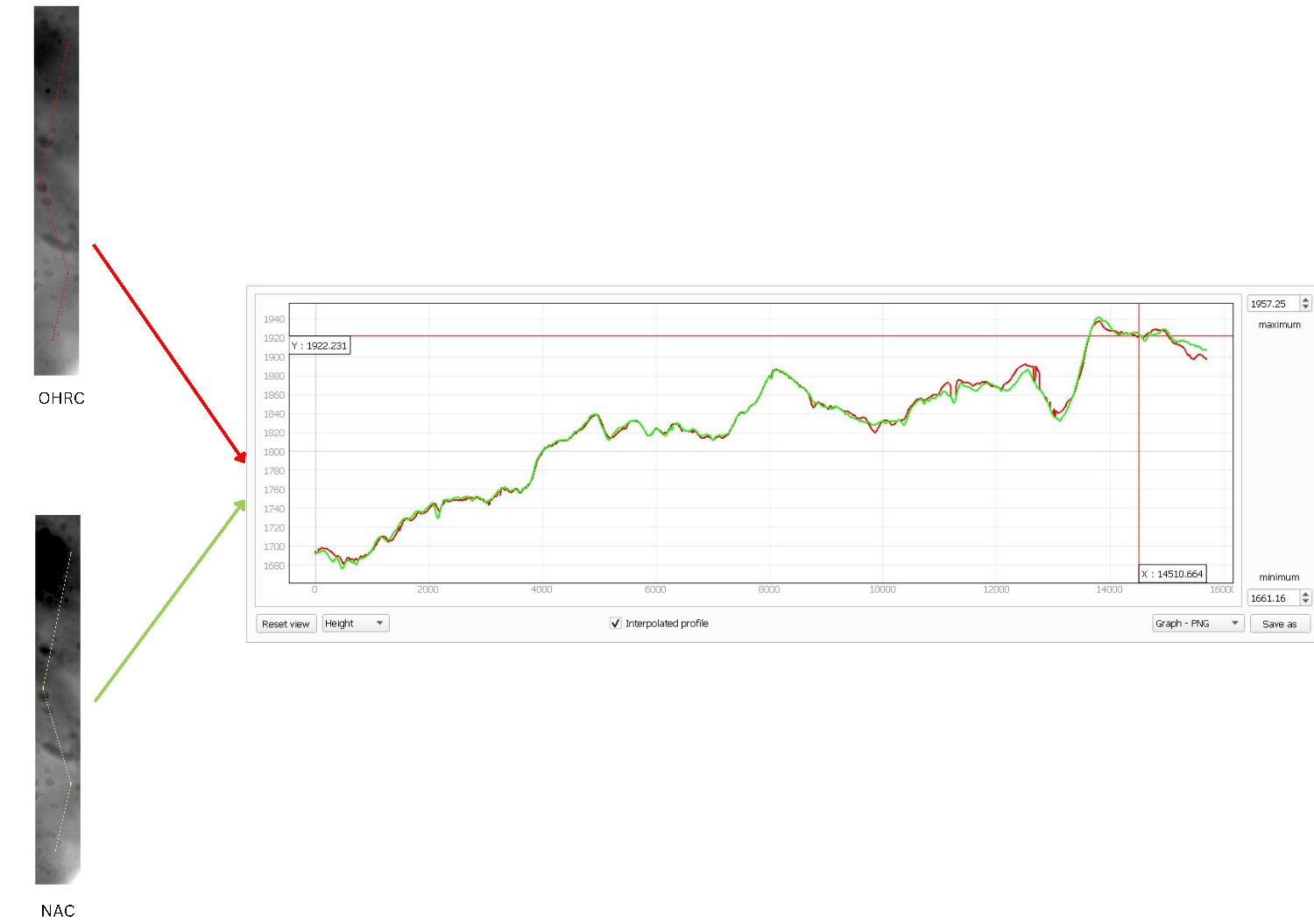}
\caption{Terrain profile comparison across Region 2. Horizontal axis: distance along profile
transect (m). Vertical axis: elevation (m). Red curve: OHRC DEM. Green curve:
NAC DTM reference.}
\label{fig:validation3}
\end{figure}

\begin{figure}[H]
\centering
\includegraphics[width=0.95\columnwidth]{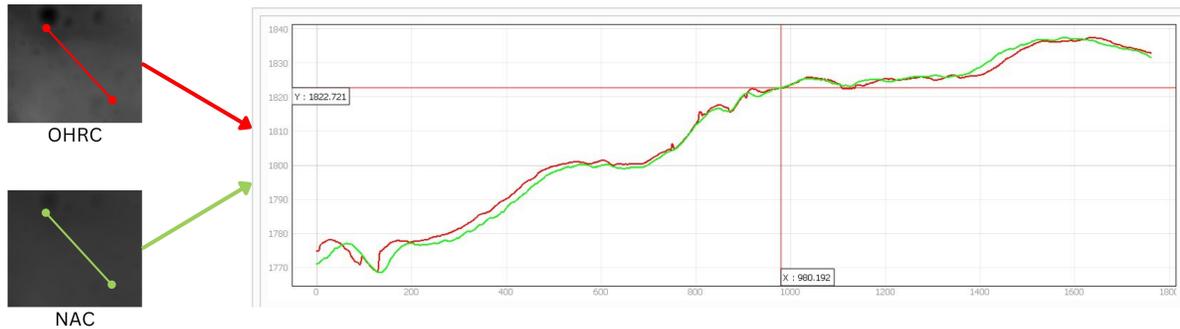}
\caption{Terrain profile comparison across a small patch of Region 2. Horizontal axis: distance along profile
transect (m). Vertical axis: elevation (m). Red curve: OHRC DEM. Green curve:
NAC DTM reference.}
\label{fig:validation4}
\end{figure}

Elevation profile comparisons for three representative regions (4,1, and 2, respectively) are shown in
Figures~\ref{fig:validation1}--\ref{fig:validation3}. Profile comparison across a small patch of region 2 is demonstrated in Figure~\ref{fig:validation4}.

\subsection{Computational Performance}

All processing was performed on a workstation equipped with an Intel Xeon Gold
6530 processor (128 logical cores) and 1\,TB of RAM. Table~\ref{tab:timing}
reports wall-clock timings for each major stage of the pipeline, measured
across the five stereo pairs. The pre-processing stages, including
data ingestion and spice kernel initialization, are collectively lightweight, completing in
approximately 1--2\,min per stereo pair. Their cost is dominated by kernel
database queries and JSON serialisation rather than computation, and scales
only weakly with image size.

Bundle adjustment, which iterates a Levenberg--Marquardt optimiser over
matched tie points until convergence or the 100-iteration limit is reached,
required approximately 10--20\,min per pair. This is consistent with reported
runtimes for comparable pushbroom stereo datasets in the planetary photogrammetry
literature \citep{beyer2018,kirk2008}.

The dominant cost is dense stereo correspondence,
which ran for approximately 10--30\,min per pair, depending primarily on image
overlap area and disparity-search range. Point cloud rasterisation required 30\,min to 2\,h per pair, with runtime governed
by the output grid resolution; finer grids at sub-30\,cm spacing incurred the
longest runtimes.

\begin{table}[H]
\centering
\caption{Approximate wall-clock runtimes per stereo pair for each pipeline
         stage. Hardware: Intel Xeon Gold 6530 $\times$128 cores, 1\,TB RAM.}
\label{tab:timing}
\renewcommand{\arraystretch}{1.25}
\begin{tabular}{lc}
\toprule
\textbf{Pipeline Stage} & \textbf{Runtime (per pair)} \\
\midrule
Data ingestion \& SPICE init     & $\sim$1--2\,min \\
Sensor model generation       & $<$1\,min \\
Bundle adjustment             & $\sim$10--20\,min \\
Dense stereo correspondence   & $\sim$10--30\,min \\
Point cloud rasterisation     & $\sim$30\,min -- 2\,h \\
ICP alignment \& offset corr. & $\sim$5--15\,min \\
\midrule
\textbf{Total (approx.)}      & $\sim$1--3\,h \\
\bottomrule
\end{tabular}
\end{table}

Stereo pair identification (convergence angle and B/H ratio computation from
image metadata) is performed as a pre-pipeline analysis step and is not
included in the above timings; it requires only seconds per candidate pair.

% =============================================================================
\section{Discussions}
\label{sec:discussion}
% =============================================================================

The results demonstrate that sub-metre lunar DEMs can be generated from
Chandrayaan-2 OHRC imagery using open-source tools, provided
that two instrument-specific obstacles are addressed. The first is pipeline
compatibility: the OHRC PDS4 data format and camera model have limited support out
of the box by existing open-source stereo processing suites, and establishing
this compatibility required developing a new import template and a CSM sensor
configuration. 

The second is stereo pair identification: because OHRC images
are not acquired as predefined stereo pairs, usable stereo combinations must be
extracted entirely from orbital geometry parameters in individual image
metadata, using B/H ratio and convergence angle as selection criteria.

A third factor affecting the absolute accuracy of the reconstructed DEMs is
the quality of the SPICE kernels currently distributed for Chandrayaan-2.
Inaccuracies in the spacecraft ephemeris or attitude kernels introduce
systematic offsets into the camera position and pointing solution, which
manifest as a constant vertical bias and, potentially, residual tilt in the
reconstructed point cloud. These effects are partially mitigated by ICP
alignment and profile-based offset correction, but cannot be fully eliminated
without more accurate kernel data. The current SPICE kernels for Chandrayaan-2
are therefore a limiting factor in the absolute vertical accuracy of any
OHRC-derived DEM, independent of the photogrammetric reconstruction quality.

The range of B/H ratios observed across Regions~1--4 (0.396--0.877) is
comparable to values reported in HiRISE stereo studies \citep{kirk2008},
confirming that the same geometric quality criteria are applicable to OHRC. For
these four regions, where B/H falls within the conventionally recommended range
of 0.3--1.0 \citep{hasegawa2000accuracy}, stereo reconstruction produced DEMs with
satisfactory spatial completeness and consistent validation metrics.

\subsection*{Region~5 as a Limiting Case}

Region~5 ($\sim\!4^{\circ}$N, $230^{\circ}$E) occupies a distinct position in
the dataset. It was selected primarily to demonstrate dataset limitations. With a B/H ratio of 1.161 and a directly measured convergence
angle of approximately $61^{\circ}$, this pair substantially exceeds the
commonly recommended B/H ceiling of 1.0 \citep{hasegawa2000accuracy}. The
high convergence angle translates directly into a large cross-view radiometric
disparity: at $\theta \approx 61^{\circ}$, surface normals that are oblique to
one camera are near-perpendicular to the other, causing marked differences in
foreshortening, shadow extent, and local brightness across the two images. This
radiometric mismatch severely degrades the ability of the block matching
algorithm to find reliable correspondences, particularly in shadow-affected
regions and on surfaces with low photometric texture.

\begin{figure}[H]
\centering
\includegraphics[width=0.95\columnwidth, height=8cm, keepaspectratio]{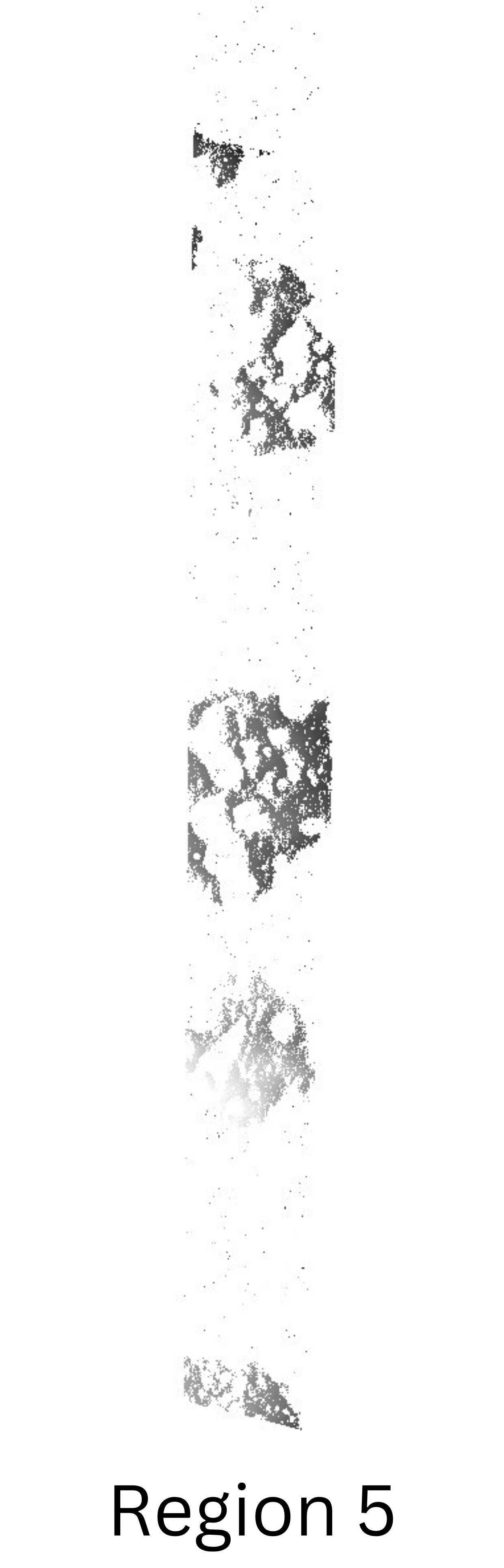}
\caption{Region 5 demonstrated with high void fraction}
\label{fig:region5}
\end{figure}

As a consequence, the DEM produced for Region~5 (Figure~\ref{fig:region5}), which was nominally the
finest-resolution product in the dataset at 24\,cm, exhibits a
substantially elevated void fraction compared to Regions~1--4. The high nominal
resolution is therefore of limited practical utility without extensive hole-filling from the NAC reference, which correspondingly reduces the fraction of
the final terrain model that is genuinely OHRC-derived.

This observation provides an empirical upper bound for OHRC stereo pair
selection. While B/H ratios up to approximately 1.16 are geometrically
processable in the sense that dense stereo reconstruction completes without
error, a practical ceiling of B/H~$\lesssim 0.9$ is recommended for OHRC to
ensure an acceptable balance between height sensitivity and spatial completeness
of the output DEM. This threshold is consistent with the theoretical prediction
from Eq.~\eqref{eq:conv_approx}: at B/H\,=\,0.9 the convergence angle is
approximately $48^{\circ}$, which represents a reasonable upper bound for
semi-global matching on high-contrast pushbroom imagery. The threshold may vary
with terrain type and solar illumination conditions, and its refinement over a
larger sample of pairs is left to future work.

Region~5 is retained as a documented negative result. Its inclusion
is intentional: it serves as a concrete illustration of the B/H upper-bound
effect and provides a reference point for future studies that may wish to
evaluate more aggressive stereo geometries with improved matching strategies
such as deep-learning-based correspondence networks.

\subsection*{Iterative Refinement via ICP-Informed Bundle Adjustment Re-Pass}

Beyond the baseline pipeline described in Section~4, an optional iterative
refinement procedure was explored as an additional exercise on selected
regions. In the standard pipeline, ICP alignment corrects the absolute
position of the reconstructed point cloud by estimating a rigid transformation
against the NAC reference. However, this correction is applied
post-reconstruction and does not propagate back into the photogrammetric
camera models. As a consequence, any residual geometric inconsistency between
the two cameras, that is small enough to survive the first bundle adjustment pass
but detectable once a reference-aligned DEM is available, remains embedded
in the camera pointing solution.

The refinement procedure closes this loop. The ICP-derived transformation
matrix, which encodes the residual offset between the initial OHRC
reconstruction and the NAC reference frame, is used to constrain a second
bundle adjustment pass. With this geometry-informed initialisation, the
optimiser converges to a camera solution that is more tightly anchored to the
absolute reference frame than was achievable with SPICE-only initialisation.
Dense stereo is then re-run under the refined camera models, and a new DEM
is gridded using an expanded interpolation search radius to recover valid
elevations in regions that were voids in the initial reconstruction. This
expanded-radius gridding is particularly effective at filling isolated no-data
pixels surrounded by valid measurements (gaps that arise not from systematic
shadow or texture failure but from marginal correspondence scores near the
stereo matching threshold).

The qualitative outcome of this procedure was a measurable improvement in
spatial completeness: regions that had previously required NAC infill were
recovered as genuinely OHRC-derived elevations, increasing the fraction of the
final DEM that carries sub-metre native resolution. This improvement was
achieved without introducing any additional external data source, relying
entirely on the self-consistency between the initial reconstruction and the NAC
reference as a geometric constraint. The full quantitative characterisation of
the accuracy improvement attributable to this refinement pass is left to future
work, where it will be evaluated systematically across a broader set of stereo
pairs.

\subsection*{Region-of-Interest Cropping for Targeted Reconstruction}

A practical consideration in processing high-resolution OHRC imagery is that
the full image swath is rarely of uniform scientific interest, and running dense
stereo over the entire frame incurs unnecessary computational cost. To address
this, a region-of-interest (ROI) approach was adopted in which only a
user-selected sub-window of each image was passed to the stereo pipeline, rather
than cropping the image files themselves.

ROI selection was performed interactively using the
graphical front-end provided by ASP, which allows the operator to inspect both
images of a stereo pair simultaneously, identify the region of overlap
containing the terrain of interest, and record the pixel coordinates of the
desired sub-window. Alternatively, ROI bounds can be derived analytically by
map-projecting the images onto a reference terrain model and identifying the
overlap footprint in geographic coordinates. The selected sub-window extents
were then passed directly to dense stereo correspondence via the
left image crop window and the right image crop window parameters,
each specified as $\langle x_{\min},\, y_{\min},\, \mathrm{width},\,
\mathrm{height} \rangle$ in pixel coordinates. This approach avoids any
modification of the source image files and preserves the full SPICE-based
camera geometry, since the crop parameters are applied internally during stereo
preprocessing.

\begin{figure}[H]
\centering
\includegraphics[width=0.95\columnwidth]{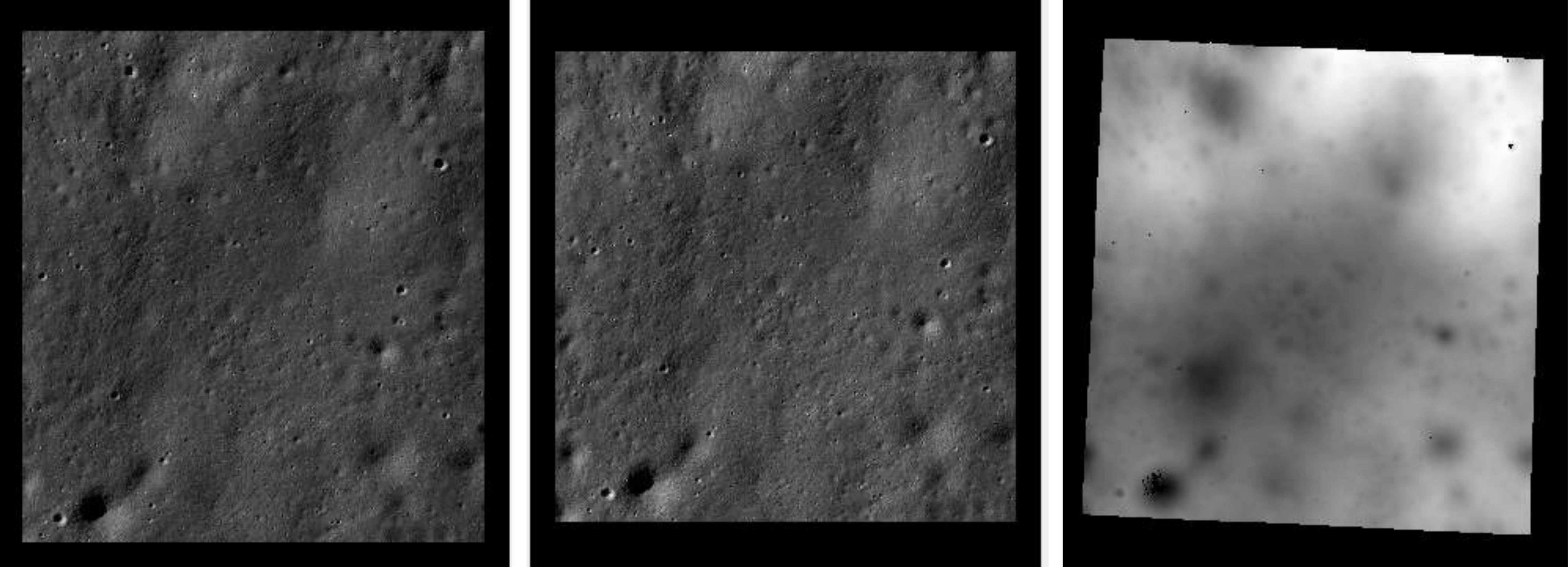}
\caption{Region-of-interest stereo reconstruction. Left and centre: the
left-camera and right-camera cropped image chips in native pixel coordinates. Right: the reconstructed
DEM for the same region.}
\label{fig:roi_dem}
\end{figure}

The resulting DEM covers only the selected sub-region, with a corresponding
reduction in processing time proportional to the area of the crop window
relative to the full image. Furthermore, the number of valid pixels obtained also increases substantially, as a result of the selected regions being appropriately illuminated. It should be noted that the reconstructed DEM may
appear geometrically offset relative to the raw cropped image chips when
displayed together: the image chips retain their native pixel coordinate system,
whereas the DEM is a georeferenced product whose spatial position is determined
by the SPICE ephemeris, CSM camera model, and bundle-adjusted camera states
established in earlier pipeline stages. The apparent tilt between the DEM and
the source imagery is therefore not a reconstruction artefact but a consequence
of correct geopositioning --- the DEM is registered to the lunar body-fixed
frame, while the image chips are not. This is illustrated in
Fig.~\ref{fig:roi_dem}, which shows the generated DEM alongside the
left and right cropped image chips for a selected patch from the images of Region 1 (~\ref{tab:dataset}).

% =============================================================================
\section{Conclusion}
% =============================================================================

This study presents a novel, reproducible open-source framework for generating and metrologically
validating sub-metre lunar DEMs from Chandrayaan-2 OHRC multi-view imagery
using photogrammetric tools. A key contribution of this work is enabling
compatibility of OHRC data with open-source pipelines through the use of a
PDS4 import template and CSM camera configuration, building upon ongoing
community-driven developments. In addition, a systematic strategy for
identifying viable stereo pairs from non-paired image archives is established
using baseline-to-height ratio and convergence angle analysis.

Stereo reconstruction is demonstrated across five geographically distributed
lunar regions, producing DEMs at spatial resolutions between 24 and 54\,cm.
Absolute elevation consistency is achieved through ICP alignment with NAC
reference terrain, followed by residual bias correction via profile-based
comparison. Continuous terrain coverage is obtained through priority-based
mosaicing. Horizontal accuracy, evaluated through planimetric feature matching
against NAC hillshade (5--10 features per region), is observed to be within
30\,cm, while vertical accuracy at native resolution yields a mean RMSE of
5.85\,m relative to NAC-derived reference data.

Future work will focus on extending spatial coverage, integrating
shape-from-shading refinement, and exploring the applicability of the
generated DEMs for landing site hazard assessment. Additionally, anticipated
updates to Chandrayaan-2 SPICE kernels, with improved ephemeris and attitude
solutions, are expected to reduce the systematic vertical offsets observed in
the current results and further improve absolute accuracy.

% =============================================================================
\section*{Acknowledgements}

The authors thank the Director, Space Applications Centre (SAC), ISRO, for
providing the opportunity to undertake this work. The authors are grateful to
the Deputy Director, SIPA, and the Head, PSPD for their encouragement and support.

The authors also thank ISRO for providing open access to the Chandrayaan-2 OHRC
imagery and SPICE kernel archive through the ISSDC PRADAN portal. The authors
also acknowledge the NASA Ames Stereo Pipeline (ASP) team and the USGS ISIS
community for their open-source software and providing the
development support for Chandrayaan-2 data, which were integral to this work.

\bibliographystyle{plainnat}
\bibliography{references}

\end{document}